\documentclass{article}
\usepackage{ijcai11}
\usepackage[boxruled,commentsnumbered,vlined]{algorithm2e}
\usepackage{graphics}
\usepackage{amssymb}
\usepackage{amsmath}
\usepackage{url}
\usepackage{multirow}
\RequirePackage{stmaryrd}
\usepackage{amsfonts} 
\usepackage{pstricks}
\usepackage{pstricks-add} 
\usepackage{fancybox} 
\newcommand{\gdi}{{\cal I}}              

\usepackage{times}

\newenvironment{definition}[1][definition]{\begin{trivlist}
\item[\hskip \labelsep {\bfseries #1}]}{\end{trivlist}}

\newcommand{\qed}{\nobreak \ifvmode \relax \else
      \ifdim\lastskip<1.5em \hskip-\lastskip
      \hskip1.5em plus0em minus0.5em \fi \nobreak
      \vrule height0.75em width0.5em depth0.25em\fi}

\title{Revisiting Numerical Pattern Mining with Formal Concept Analysis}

\author{
Mehdi Kaytoue$^1$ , Sergei O. Kuznetsov$^2$, and Amedeo Napoli$^1$\\
$^1$LORIA (CNRS -- INRIA Nancy Grand Est -- Nancy Universit\'e)\\
Campus Scientifique, B.P. 70239 -- Vand{\oe}uvre-l\`{e}s-Nancy -- France\\
$^2$Higher School of Economics -- State University\\Pokrovskiy Bd. 11 -- 109028 Moscow -- Russia  \\
\textit{kaytouem@loria.fr, skuznetsov@hse.ru, napoli@loria.fr}
}

\begin{document}

\maketitle

\begin{abstract}
We investigate the problem of mining numerical data with Formal Concept Analysis. The usual way is to use a scaling procedure --transforming numerical attributes into binary ones-- leading either to a loss of information or of efficiency, in
particular w.r.t. the volume of extracted patterns.
By contrast, we propose to directly work on numerical data in a more
precise and efficient way. For that, the notions of closed
patterns, generators and equivalent classes are revisited in the
numerical context. Moreover, two algorithms are proposed and
tested in an evaluation involving real-world data, showing the
quality of the present approach. 
\end{abstract}

\section{Introduction}

In this paper, we investigate the problem of mining numerical data.
This problem arises in many practical situations, e.g. analysis of gene and
transcriptomic data in biology, soil characteristics and land
occupation in agronomy, demographic data in economics, temperatures in
climate analysis, etc.
We introduce an original framework for mining numerical
data based on advances in itemset mining and in Formal Concept
Analysis (FCA,~\cite{GanterWille99}), respectively condensed representations of itemsets and pattern structures in FCA~\cite{GanterK01}.
The mining of frequent itemsets in binary data, considering a set of
objects and a set of associated attributes or items, is studied for a
long time and usually involves the so-called ``pattern flooding''
problem \cite{BastideTPSL00}.
A way of dealing with pattern flooding is to search for equivalence
classes of itemsets, i.e. itemsets shared by the same set of objects
(or having the same image).
For an equivalence class, there is one maximal itemset, which
corresponds to a ``closed set'', and possibly several minimal elements
corresponding to ``generators'' (or ``key itemsets'').
From these elements, families of association rules can be extracted.
These itemsets are also related to FCA , where a
concept lattice is built from a binary context and where formal concepts are closed sets of objects and attributes.

The present work is rooted both in FCA and pattern mining with the
objective of extracting interval patterns from numerical data. 
Our approach is based on ``pattern structures'' where
complex descriptions can be associated with objects. 
In~\cite{IS} we introduced closed interval patterns in the context of gene expression data mining. Intuitively, an interval pattern is a vector of intervals, each dimension corresponding to a range of values of a given attribute ; it is closed when composed of the smallest intervals characterizing a same set of objects.

In the present paper, we complete and extend this first attempt.
Considering numerical data, some general characteristics of
equivalence classes remain, e.g. one maximal element which is a
closed pattern and possibly several generators which are minimal
patterns w.r.t. a subsumption relation defined on patterns.
We show that directly extracting patterns data from numerical is more
efficient using pattern structures than working on binary data with
associated scaling procedures.
We also provide a semantics to interval patterns in the Euclidean space, design and experiment algorithms to extract frequent closed interval patterns and their generators.

The problem of mining patterns in numerical data is usually referred as quantitative itemset/association rule mining~\cite{SrikantA96}. Generally, an appropriate discretization splits attribute ranges into intervals maximizing some interest functions, e.g. support, confidence.  However, none of these works covers the notion of equivalence classes, closed patterns, and generators, and this is one of the originality of the present paper.


The plan of the paper is as follows. Firstly, we introduce the problem of mining numerical data and interval patterns. Then, we recall the basics of FCA and interordinal scaling. We pose a number of questions that we propose to answer using our framework of interval pattern structures dealing with numerical
data. We then detail two original algorithms for extracting closed interval
patterns and their generators.  These algorithms are evaluated in the last section on real-world data. Finally, we end the paper in discussing related work and giving
perspectives to the present research work. As a complement, an extended version of this paper is given in~\cite{RR2}, providing algorithms pseudo-code and a longer discussion on the usefulness of interval patterns  in classification problems and privacy preserving data-mining.

\section{Problem definition}
We propose a definition of interval patterns for numerical data. Intuitively, each object of a numerical dataset corresponds to a vector of numbers, where each dimension stands for an attribute. Accordingly, an interval pattern is a vector of intervals, and each dimension describes the range of a numerical attribute. We only consider finite intervals and that the set of attributes/dimensions is assumed to bed (canonically) ordered.

\begin{definition}[Numerical dataset.]
A numerical dataset is given by a set of objects $G$, a set of numerical attributes $M$, where the range of $m \in M$ is a finite set noted $W_m$. $m(g)=w$ means that $w$ is the value of attribute $m$ for object $g$.
\end{definition}
\begin{table}[h]  \centering
\begin{scriptsize}
\vspace{-0.4cm}
\begin{tabular}{|l||l|l|l|}
\hline
~ &$m_1$&$m_2$&$m_3$\\
\hline\hline
$g_1$ & 5 & 7 & 6 \\
$g_2$ & 6 & 8 & 4 \\
$g_3$ & 4 & 8 & 5 \\
$g_4$ & 4 & 9 & 8 \\
$g_5$ & 5 & 8 & 5 \\
\hline
\end{tabular}
\end{scriptsize}
\caption{A numerical dataset.}
\label{tab:data}
\end{table}
\vspace{-0.4cm}
\begin{definition}[Interval pattern and support.]
In a numerical dataset, an interval pattern is a vector of intervals $d = \langle [a_i, b_i] \rangle_{i \in \{1, ..., |M|\}}$ where $a_i,b_i \in W_{m_i}$, and each dimension corresponds to an attribute following a canonical order on vector dimensions, and $|M|$ denotes the number of attributes.  An object $g$ is in the image of an interval pattern $\langle [a_i, b_i] \rangle_{i \in \{1, ..., |M|\}}$ when $m_i(g) \in [a_i, b_i]$, $\forall i \in \{1,...,|M|\}$. The support of $d$, denoted by $sup(d)$, is the cardinality of the image of $d$.
\end{definition}
\noindent\textit{Running example.} Table~\ref{tab:data} is a numerical dataset with objects in $G = \{g_1, ..., g_5\}$, attributes in $ M = \{ m_1, m_2, m_3 \}$. The range of $m_1$ is $W_{m_1} = \{4,5,6\}$, and we have $m_1(g_1)= 5$. 
Here, we do not consider either missing values or multiple values for an attribute. $\langle [5,6],[7,8],[4,6] \rangle$ is an interval pattern in Table~\ref{tab:data}, with image $\{g_1,g_2,g_5\}$ and support $3$. 

\begin{definition}[Interval pattern search space.]
Given a set of attributes $M = \{ m_i\}_{i \in \{1,|M|\}}$, 
the search space of interval patterns is the set $D$ of all interval vectors $\langle [a_i, b_i] \rangle_{i \in \{1,...,|M|\}}$, with $a_i,b_i \in W_{m_i}$.
The size of the search space is given by  $$ |D| = \prod_{i \in \{1,...,|M|\}}  (|W_{m_i}| \times  (|W_{m_i}| + 1))/2$$
\end{definition}

\noindent\textit{Example.} All possible intervals for $m_1$ are in $\{[4,4],[5,5],[6,6],[4,5],[5,6]$, $[4,6]\}$. Considering also attributes $m_2$ and $m_3$, we have  $6 \times 6 \times 10 = 360$ patterns.

\medskip

The classical problem of ``pattern flooding'' in data-mining is even worst for numerical data. Indeed, with three attributes, there are only $2^3 = 8$ possible itemsets, compared to the 360 interval patterns in the above example with same number of attributes. A solution widely investigated in itemset-mining for minimizing the effect of pattern flooding relies on  condensed representations including closed itemsets and generators~\cite{BastideTPSL00}. By contrast, the analysis of numerical datasets  can be considered within the formal concept analysis framework (FCA)~\cite{GanterWille99}, which is closely related to itemset-mining~\cite{StummeTBPL02}. Accordingly, we are interested in adapting the notions of (frequent) closed itemsets and their generators  to interval patterns within the FCA framework, and in providing an appropriate semantics to these patterns.

\section{Interval patterns in FCA}
\cite{GanterWille99} define a discretization procedure, called \textit{interordinal scaling}, transforming numerical data into binary data that encodes any interval of values from a numerical dataset. We recall here the basics on FCA and interordinal scaling.

\begin{table*} \centering
\begin{center}
\begin{tabular}{|l||l|l|l|l|l|l||l|l|l|l|l|l||l|l|l|l|l|l|l|l|}
\hline
& \rotatebox{90}{$m_1\leq 4~$} & \rotatebox{90}{$m_1\leq 5~$} & \rotatebox{90}{$m_1\leq 6~$} & \rotatebox{90}{$m_1\geq 4~$} & \rotatebox{90}{$m_1\geq 5~$} & \rotatebox{90}{$m_1\geq 6~$} & \rotatebox{90}{$m_2\leq7~$} & \rotatebox{90}{$m_2\leq8~$} & \rotatebox{90}{$m_2\leq9~$} & \rotatebox{90}{$m_2\geq7~$} & \rotatebox{90}{$m_2\geq8~$} & \rotatebox{90}{$m_2\geq9~$} & \rotatebox{90}{$m_3\leq4~$} & \rotatebox{90}{$m_3\leq5~$} & \rotatebox{90}{$m_3\leq6~$} & \rotatebox{90}{$m_3\leq8~$} & \rotatebox{90}{$m_3\geq4~$} & \rotatebox{90}{$m_3\geq5~$} & \rotatebox{90}{$m_3\geq6~$} & \rotatebox{90}{$m_3\geq8~$}\\
\hline
\hline
 $g_1$ & & $\times$ & $\times$ & $\times$ & $\times$ &  & $\times$ & $\times$ & $\times$ & $\times$ &  &  &  &  & $\times$ & $\times$ & $\times$ & $\times$ & $\times$ & \\
$g_2$ & & & $\times$ & $\times$ & $\times$ & $\times$ &  & $\times$ & $\times$ & $\times$ & $\times$ &  & $\times$ & $\times$ & $\times$ & $\times$ & $\times$ &  &  &  \\
$g_3$ & $\times$ & $\times$ & $\times$ & $\times$ &  &  &  & $\times$ & $\times$ & $\times$ & $\times$ &  &  & $\times$ & $\times$ & $\times$ & $\times$ & $\times$ &  & \\
$g_4$ & $\times$ & $\times$ & $\times$ & $\times$ &  &  &  &  & $\times$ & $\times$ & $\times$ & $\times$ &  &  &  & $\times$ & $\times$ & $\times$ & $\times$ & $\times$\\
$g_5$ & & $\times$ & $\times$ & $\times$ & $\times$ &  &  & $\times$ & $\times$ & $\times$ & $\times$ &  &  & $\times$ & $\times$ & $\times$ & $\times$ & $\times$ &  & \\
\hline
\end{tabular} 
\end{center}
\vspace{-0.3cm}
\caption{Interordinally scaled context encoding the numerical dataset from Table~\ref{tab:data}.}
\vspace{-0.4cm}
\label{tab:interordinally-scaled}
\end{table*}

\subsection{Formal concept analysis}
FCA starts with a formal context $(G,N,I)$ where
$G$ denotes a set of objects, $N$ a set of attributes, or items,
and $I \subseteq G \times N$ a binary relation between $G$ and $N$.
The statement $(g,n) \in I$, or $gIn$, means: ``the object $g$ has
attribute $n$''.
Two operators~$(\cdot)'$ define a Galois connection
between the powersets $(\mathfrak P (G), \subseteq)$ and $(\mathfrak P (N), \subseteq)$,
with $A \subseteq G$ and $B \subseteq N$:
$A'=\{n \in N \mid \forall g \in A : gIn \}$ and $B'=\{g \in G \mid \forall n \in B : gIn\}$.
A pair $(A,B)$, such that $A' = B$
and \mbox{$B' = A$}, is called a \textit{(formal) concept}, while
$A$ is called the $extent$ and $B$ the $intent$ of the concept $(A,B)$.

From an itemset-mining point of view, concept intents correspond to closed itemsets, since $(.)''$ is a closure operator. An equivalence class is a set of itemsets with same closure (and  same image). For any subset $B \subseteq N$, $B''$ is the largest itemset w.r.t. set inclusion in its equivalence class. Dually, generators are the smallest itemsets w.r.t. set inclusion in an equivalence class. Precisely, $B \subseteq N$ is closed iff $\nexists C$ such as $B \subset C$ with $C' = B'$ ; $B \subseteq N$ is a generator iff $\nexists C \subset B$ with $C' = B'$.
\subsection{Interordinal scaling}
Given a numerical attribute $m$ with range $W_m$, Interordinal Scaling builds a binary table with $2 \times |W_m|$ binary attributes. They are denoted by ``$m \leq w$'' and ``$m \geq w$'', $\forall w \in W_m$, and called IS-items. An object $g$  has an IS-item ``$m \leq w$'' (resp. ``$m \geq w$'') iff $m(g) \leq w$ (resp. $m(g) \geq w)$. Applying this scaling to our example gives Table~\ref{tab:interordinally-scaled}. It is possible to apply classical mining algorithms to process this table for extracting itemsets composed of IS-items.  These itemsets are called IS-itemsets in the following. 

IS-itemsets can be turned into interval patterns, since an IS-item gives a constraint on the range $W_m$ of an attribute $m$. For example, the IS-itemset $\{m_1\leq 5, m_1\leq 6, m_1\geq 4, m_2\leq 9, m_2\geq 7\}$ corresponds to the interval pattern $\langle [4,5],[7,9], [4,8] \rangle$. We have here the interval $[4,8]$ for attribute $m_3$: $[4,8]$ covers the whole range of $m_3$ since no constraint is given for $m_3$. 

Therefore, mining interval patterns can be considered with a scaling of numerical data. However, this scaling produces a very important number of binary attributes compared to the original ones. Hence, when original data are very large, the size of the resulting formal context involves hard computations. Accordingly, this raises the following questions:

(i) Can we avoid scaling and directly work on numerical data instead of searching for IS-itemsets? (ii) Can we adapt the notions of condensed representations such as closed patterns and generators for numerical data, and efficiently compute those patterns? (iii) What would be the semantics that could be provided to closed patterns and generators?
\section{Revisiting numerical pattern mining}
In this section, we answer those questions. First, we show that a closure operator can be defined for interval patterns based on their image. Then, we provide interval patterns with an appropriate semantics for defining the notion of equivalence classes of patterns, closed and generator patterns. After discussing why working with interordinal scaling is not acceptable thanks to the  semantics of interval patterns, we propose two efficient algorithms for mining closed interval patterns and generators. Experiments follow in Section 5.
\subsection{A closure operator for interval patterns}
We introduce  the formalism of pattern structures \cite{GanterK01}, an extension of formal contexts for dealing with complex data in FCA. It defines a closure operator for a partially ordered set of object descriptions called patterns.

Formally, let $G$ be a set of objects, $(D, \sqcap)$ be a semi-lattice of object descriptions, and $\delta : G \rightarrow D$ be a mapping: $(G, (D,\sqcap), \delta)$ is called a \textit{pattern structure}. Elements of $D$ are called \textit{patterns}, and are ordered as follows $c \sqcap d = c \iff c \sqsubseteq d$. Intuitively, objects in $G$ have descriptions in $(D,\sqcap)$. For example, $g_1$ in Table 1 has description $\langle [5,5],[7,7],[6,6] \rangle$ where $D$ is the set of all possible interval patterns ordered with $\sqcap$, as made precise below. Consider the two operators $(.)^\square$ defined as follows, with  $A \subseteq G$ and $d  \in (D,\sqcap)$
\begin{center}
$d^\square=\{g \in G | d \sqsubseteq \delta (g) \}~~~~~~~~~~~A^\square =  \bigsqcap _{g \in A} \delta (g)$
\end{center}
These operators form a Galois connection between \mbox{$(\mathfrak P (G),\subseteq)$} and $(D, \sqsubseteq)$. $(.)^{\square\square}$ is a closure operator, meaning that any pattern $d$ such as $d = d^{\square\square}$ is closed.

\medskip

\noindent\textbf{Interval pattern structures.} This general closure operator can be used for interval patterns. Indeed, interval patterns can be ordered within a meet-semi-lattice when the infimum is defined as follows. Let $c = \langle [a_i, b_i] \rangle_{i \in \{1,...,|M|\}}$, and $d = \langle [e_i, f_i] \rangle_{i \in \{1,...,|M|\}}$ two intervals patterns. Their infimum is given by $c \sqcap d = \langle [min(a_i,e_i), max (b_i,f_i)] \rangle_{i \in \{1,...,|M|\}}$. The ordering relation induced by this definition is: $c \sqsubseteq d \iff [e_i, f_i] \subseteq [a_i, b_i],~\forall i\in \{1,...,|M|\}$.

Consider now a numerical dataset, e.g. Table~\ref{tab:data}. $(D,\sqsubseteq)$ is the finite ordered set of all interval patterns. $\delta(g) \in D$ is the pattern associated to an object $g \in G$. Then:
\begin{equation*}
\begin{array}{l}
\langle [5,6], [7,8], [4,8] \rangle ^\square{}=\{g \in G | \langle [5,6], [7,8], [4,8] \rangle \sqsubseteq \delta (g) \} \\
~~~~~~~~~~~~~~~~~~~~~~~~~~~~~~~~~~=\{g_1, g_2, g_5\}\\
\{g_1, g_2, g_5\}^\square = \delta(g_1) \sqcap \delta(g_2) \sqcap \delta(g_3)\\
~~~~~~~~~~~~~~~~~~~~~=\langle [5,6], [7,8], [4,6] \rangle 
\end{array}
\end{equation*}
This means that $\langle [5,6], [7,8], [4,8] \rangle$ is not a closed interval pattern, its closure being $\langle [5,6], [7,8], [4,6] \rangle$. 

\subsection{Semantics}
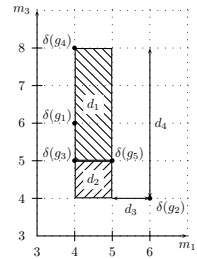
\begin{figure}[h]
\begin{minipage}{0.40\linewidth}\centering
\scalebox{0.5}{
\begin{pspicture}(0,-0.2)(1,6)
   \psaxes[Ox=3,Oy=3]{->}(0,0)(0,0)(4,6)
   \psgrid[griddots=5, subgriddiv=0, gridlabels=0pt](0,0)(4,6)

	\psdots*(1,3)(3,1)(1,2)(1,5)(2,2)
	\uput[l](1.1,3.2){$\delta(g_1)$}
	\uput[r](3,0.75){$\delta(g_2)$}
	\uput[l](1.1,2.2){$\delta(g_3)$}
	\uput[l](1.1,5.2){$\delta(g_4)$}
	\uput[r](1.9,2.2){$\delta(g_5)$}
	   
	\psframe[fillstyle=vlines](1,2)(2,5)
	\psframe[fillstyle=hlines](1,1)(2,2)
	
	\psline{<->}(2,1)(3,1)
	\psline{<->}(3,1)(3,5)
	
	\uput[d](4,0){$m_1$}	
	\uput[l](0,6){$m_3$}	

	\rput(1.5,3.5){\psframebox*[fillstyle=solid,fillcolor=gray]{$d_1$}}
	\rput(1.5,1.5){\psframebox*[fillstyle=solid,fillcolor=lightgray]{$d_2$}}		
	\rput(1.5,3.5){\psframebox*[fillstyle=solid,fillcolor=white]{$d_1$}}
	\rput(1.5,1.5){\psframebox*[fillstyle=solid,fillcolor=white]{$d_2$}}	
	\rput(2.5,0.7){\psframebox*{$d_3$}}	
	\rput(3.3, 3){\psframebox*{$d_4$}}	
	
\end{pspicture}
}
\end{minipage}
\hfill
\begin{minipage}[b]{0.49\linewidth}\centering
\begin{tabular}{l}
~~~~ $d_1 = \langle [4,5],[5,8]\rangle$ \\ 
~~~~ $d_1^\square = \{g_1, g_3,g_4, g_5\}$ \\ 
~~~~ $d_2 = \langle [4,5],[4,5]\rangle$ \\ 
~~~~ $d_2^\square = \{g_3, g_5\}$ \\ 
~~~~ $d_3 = \langle [5,6],[4,4]\rangle$ \\ 
~~~~ $d_3^\square = \{g_2\}$ \\ 
~~~~ $d_4 = \langle [6,6],[4,8]\rangle$  \\ 
~~~~ $d_4^\square = \{g_2\}$
\end{tabular} 
\end{minipage}
\caption{Interval patterns in the Euclidean space.}
\label{eucli}
\end{figure}

An interval pattern $d$ is a $|M|$-dimensional vector of intervals and can be represented by a hyper-rectangle (or rectangle for short) in Euclidean space $\mathbb{R}^{|M|}$, whose sides are parallel to the coordinate axes. This  geometrical representation provides a semantics for interval patterns. In formal terms, an interpretation is given by $\gdi = (\mathbb{R}^{|M|},(.)^\gdi)$ where $\mathbb{R}^{|M|}$ is the interpretation domain, and \mbox{$(.)^\gdi:D \rightarrow \mathbb{R}^{|M|}$}  the interpretation function. Figure~\ref{eucli} gives four interval pattern representations in $\mathbb{R}^2$, with only attributes $m_1$ and $m_3$ of our example. The image of $d_1$ is given by all objects $g$ whose description $\delta(g)$ is included in the rectangle associated with $d_1$, i.e. the set $\{g_1, g_3,g_4, g_5\}$. We can interpret the closure operator $(.)^{\square\square}$ according to this semantics. The first operator $(.)^\square$ applies to a rectangle and returns the set of objects whose description is included in this rectangle. The second operator $(.)^\square$ applies to a set of objects and returns the smallest rectangle that contains their descriptions, i.e. the convex hull of their corresponding descriptions.

\vspace{-.5cm}

\subsection{Closed interval patterns and generators}

Now, we can revisit the notion of equivalence classes of itemsets as introduced in~\cite{BastideTPSL00}: an equivalence class of interval patterns is a set of rectangles containing the same object descriptions (based on all rectangles in the search space as given in Section 2). This enables to define the notions of (frequent) closed interval patterns ((F)CIP) and (frequent) interval pattern generators ((F)IPG), adapted itemisets.
\begin{definition}[Equivalence class.]
Two interval patterns $c$ and $d$  with same image are equivalent, i.e. $c^\square = d^\square$ and we write $c \cong d$. $\cong$ is an equivalence relation, i.e. reflexive, transitive and symmetric. The set of patterns equivalent to a pattern $d$ is denoted by \mbox{$[d] = \{c | c \cong d\}$} and called the equivalence class of $d$.
\end{definition}
\vspace{-.25cm}
\begin{definition}[Closed interval pattern (CIP).]
A pattern $d$ is closed if there does not exist any pattern $e$ such as $d \sqsubseteq e$ with  $d \cong e$.
\end{definition}
\vspace{-.3cm}
\begin{definition}[Interval pattern generator (IPG).]
A pattern $d$ is a generator if there does not exist a pattern $e$ such as $e \sqsubseteq d$ with  $d \cong e$.
\end{definition}
\vspace{-.3cm}
\begin{definition}[Frequent Interval pattern.]
A pattern $d$ is frequent if its image has a higher cardinality than 
a given minimal support threshold $minSup$.
\end{definition}

We illustrate these definitions with two dimensional interval patterns, and their representation in Figure~\ref{eucli}, i.e. considering attributes $m_1$ and $m_3$ only. $\langle [4,5],[6,8] \rangle \cong \langle [4,6],[6,8] \rangle$ with image $\{g_1, g_4\}$. $\langle [4,6],[6,8] \rangle$ is not closed as $\langle [4,6],[6,8] \rangle \sqsubseteq \langle [4,5],[6,8] \rangle$, these two patterns having same image, i.e. $\{g_1, g_3,g_4,g_5\}$. $\langle [4,5],[5,8] \rangle $ is closed. $\langle [4,6],[5,8] \rangle$ and $\langle [4,5],[4,8] \rangle$ are generators in the class of the closed interval pattern $d_1 = \langle [4,5],[5,8] \rangle$ with image $\{g_1,g_3,g_4,g_5\}$. Among the four patterns in Figure~\ref{eucli}, $d_1$ is the only frequent interval pattern  with $minSup=3$. 

Based on the above semantics, an equivalence class is a set of rectangles containing the same set of object descriptions, with a  (unique) closed pattern corresponding to the smallest rectangle, and one or several generator(s) corresponding to the largest rectangle(s).

These definitions are counter-intuitive w.r.t. itemsets: the smallest rectangles subsume the largest ones. This is due to the definition of infimum as set intersection for itemsets while this is convex hull for intervals, which behaves dually as a supremum. 

\subsection{IS-itemsets versus interval patterns}
Interordinal scaling allows to build binary data encoding all interval of values from a numerical dataset. Therefore, one may attempt to mine closed itemsets and generators in these data with existing data-mining algorithms. Here we show why this should be avoided.

\textbf{Local redundancy of IS-itemsets.} 
Extracting all IS-itemsets in our example (from Table~\ref{tab:interordinally-scaled}) gives $31,487$ IS-itemsets. This is surprising since there are at most $360$ possible interval patterns.  In fact, many IS-itemsets are locally redundant. For example, \mbox{$\{m_1\leq 5\}$} and $\{m_1 \leq 5, m_1 \leq 6\}$  both correspond to interval pattern $ \langle [4,5], [7,9], [4,8] \rangle$: the constraint $m_1 \leq 6$ is redundant w.r.t. $m_1 \leq 5$ on the set of values $W_{m_1}$. Hence there is no 1-1-correspondence between IS-itemsets and interval patterns. It can be shown that there is a 1-1-correspondence only between closed IS-itemsets and CIP~\cite{IS}. Later we show that local redundancy of IS-itemsets makes the computation of closed sets very hard.


\textbf{Global redundancy of IS-itemset generators.}
Since IS-itemset generators are the smallest itemsets, they do not suffer of local redundancy. However, we can remark another kind of redundancy, called global redundancy: it happens that two different and incomparable IS-itemset generators correspond to two different interval pattern generators, but one subsuming the other. In Table~\ref{tab:interordinally-scaled}, both IS-itemsets $N_1 = \{m_1 \leq 4, m_3 \leq 5\}$ and $N_2 =\{m_1 \leq 4, m_3 \leq 6\}$ have the same image $\{g_3\}$ and are generators, i.e. there does not exist a smaller itemset of these itemsets with same image. However, their corresponding interval pattern are respectively $c = \langle [4,4],[7,9],[4,5]\rangle$ and $d = \langle [4,4],[7,9],[4,6]\rangle$ and we have $d \sqsubseteq c$, while $c^\square = d^\square$, hence $c$ is not an interval pattern generator.



\subsection{Algorithms}\label{fcip}
We detail a depth-first enumeration of interval patterns, starting with the most frequent one. Based on this enumeration, we design the algorithms \textit{MinIntChange} and \textit{MinIntChangeG} for extracting respectively frequent closed interval patterns (FCIP) and frequent interval pattern generators (FIPG).

\medskip

\noindent\textbf{Interval pattern enumeration.} Consider firstly one numerical attribute of the example, say $m_1$. The semi-lattice of intervals $(D_{m_1}, \sqcap)$ is composed of all possible intervals with bounds in $W_{m_1}$ and is ordered by the relation $\sqsubseteq$. The unique smallest element w.r.t. $\sqsubseteq$ is the interval with maximal size, i.e. $[4,6] = [min(W_{m_1}),max(W_{m_1})]$ and maximal frequency (here $5$). The basic idea of pattern generation lies in \textit{minimal changes} for generating the direct subsumers of a given pattern. For example, two minimal changes can be applied to $[4,6]$.  The first consists in replacing the right bound with the value of $W_{m_1}$ immediately lower that $6$, i.e. $5$, for generating the interval $[4,5]$. The second consists in repeating the same operation for the left bound, generating the interval $[5,6]$. Repeating these two operations allows to enumerate all elements of $(D_{m_1}, \sqcap)$. A right minimal change is defined formally as, given $a,b,v\in W_m$, $a \not = b$, $mcr([a,b])= [a,v]$ with $v < b$ and $\nexists  w\in W_m$ s.t. $v < w < b$ while a left minimal change $mcl([a,b])$ is formally defined dually. Minimal changes give direct next subsumers and implies a monotonicity property of frequency, i.e. support of $[a,v]$ is less than or equal to support of $[a,b]$. To avoid generating several times the same pattern, a lectic order on changes, or equivalently on patterns, is defined. After a right change, one can apply either a right or left change; after a left change one can apply only a left change. Figure~\ref{traversal} shows the depth-first traversal (numbered arrows) of diagram of $(D_{m_1}, \sqcap)$. Backtrack occurs when an interval of the form $[w,w]$ is reached ($w \in W_{m_1}$), or no more change can be applied. Each minimal change can be interpreted in term of an IS-item. For example, if $[a,b]$ corresponds to the IS-itemsets $\{m \geq a, m \leq b\}$ then $mcr([a,b])=[a,v]$ corresponds to $\{m \geq a, m \leq b, m \leq v\}$, i.e. adding $m \leq v$ to the original IS-itemset. The same applies dually to left minimal changes. 
These IS-items characterizing minimal changes are drawn on Figure~\ref{traversal}. This figure accordingly represents a prefix-tree, factoring out the effort to process common prefixes or minimal changes, and avoiding redundancy problems inherent in interordinal scaling. The generalization to several attributes is straightforward. A lectic order is classically defined on numerical attributes as a lexicographic order, e.g. $m_1 < m_2 < m_3$. Then changes are applied as explained above for all attributes respecting this order, e.g. after applying a change to attribute $m_2$, one cannot apply a change to attribute $m_1$.
\begin{figure}[h]\centering
\scalebox{0.5}{
\begin{pspicture}(6,-.2)(0,3.5)
\cnodeput(0,0){A}{[4,4]}
\cnodeput(3,0){B}{[5,5]}
\cnodeput(6,0){C}{[6,6]}
\cnodeput(1.5,1.5){D}{[4,5]}
\cnodeput(4.5,1.5){E}{[5,6]}
\cnodeput(3,3){F}{[4,6]}
\psset{nodesep=3pt}
\ncarc{->}{A}{D}\Aput{\underline{$m_1\leq 4$}~3}
\ncarc{->}{D}{A}\Aput{2}
\ncarc{->}{B}{D}\Aput{5}
\ncarc{->}{D}{B}\Aput{4~\underline{$m_1\geq5$}}
\ncarc{->}{C}{E}\Aput{9}
\ncarc{->}{E}{C}\Aput{8~\underline{$m_1\geq6$}}
\ncarc{->}{F}{D}\Aput{1}
\ncarc{->}{D}{F}\Aput{\underline{$m_1\leq 5$}~6}
\ncarc{->}{E}{F}\Aput{10}
\ncarc{->}{F}{E}\Aput{7~\underline{$m_1\geq5$}}
\end{pspicture}
}
\caption{Depth-first traversal of ($D_{m_1},\sqcap$).}

\label{traversal}
\end{figure}

\noindent\textbf{Extracting FCIP with MintIntChange.} The pattern enumeration starts with the minimal pattern w.r.t $\sqsubseteq$ and generates its direct subsumers with lower or equal support. The next problem now is that minimal changes do not necessarily generate patterns with strictly smaller support. Therefore,  we should apply changes until a pattern with different support is generated to identify a closed interval pattern (FCIP) but this would not be efficient. We adopt the idea of the algorithm \textit{CloseByOne}~\cite{KuznetsovO02}: before applying a minimal change, the closure operator $(.)^{\square\square}$ is applied to the current pattern, allowing to skip all equivalent patterns. Indeed, the minimal pattern $d$ w.r.t. $\sqsubseteq$ is closed as it is given by $d = G^\square$. Applying a minimal change returns a pattern $c$ with strictly smaller support, since $d \sqsubseteq c$ and $d$ is closed. If $c$ is frequent, we can continue, apply the closure operator and next changes in lectic order, allowing to completely enumerate all FCIP. Since a FCIP may have several different associated generators, it can be generated several times. Still following the idea of \textit{CloseByOne}, a canonicity test can be defined according to lectic order minimal changes.

Consider a pattern $d$ generated by a change at attribute $m_j \in M$. Its closure is given by $d^{\square\square}$. If $d^{\square\square}$ differs from $d$ for some attributes $m_h \in M$ such as $m_h < m_j$, then $d^{\square\square}$ has already been generated: it is not canonically generated, hence the algorithms backtracks.

\medskip

\noindent\textit{Example.} We start from the minimal pattern $d = \langle [4,6],[7,9],[4,8]\rangle$. The first minimal change in lectic order is a right change on attribute $m_1$. We obtain pattern  $c = \langle [4,5],[7,9],[4,8]\rangle$, and obviously $d \sqsubseteq c$. However, $c^{\square\square} = \langle [4,5],[7,9],[5,8]\rangle$, hence $c$ is not closed. $c^{\square\square}$ is stored as FCIP and next changes will be applied to it. 

Now consider the pattern obtained by minimal change on left border for attribute $m_3$, i.e. $e= \langle [4,6],[7,9],[5,8] \rangle$. We have $e^{\square\square} =\langle [4,5],[7,9],[5,8] \rangle$. $e$ and $e^{\square\square}$ differ for attribute $m_1$, but $e$ has been generated from a change on $m_3$. Since $m_1 < m_3$, $e^{\square\square}$ is not canonical and has already been generated (previous example), hence the algorithm  backtracks.
%
%

\medskip

\noindent\textbf{Extracting FIPG with MintIntChangeG.} We now adapt \textit{MinIntChange} to extract FIPG, following a well-known principle in itemset-mining algorithms~\cite{CaldersG05}. For any FCIP $d$, a minimal change implies that the support of the resulting pattern $c$ is strictly smaller than the support of $d$. Therefore, $c$ is a good generator candidate of the next FCIP. Accordingly, at each step of the depth-first enumeration a FIPG candidate $c$ is generated from the previous one $b$, by applying a minimal change characterized by $b^{\square\square}$. Then, each candidate $c$ has to be checked whether it is a generator or not. We know that the candidate has no subsumers in its branch with same support. However, it could exist a branch with another FIPG $e$ with same image and resulting from less changes. Considering the lectic order on minimal changes, we use a reverse traversal of the tree (see Figure~\ref{traversal}: 7,8,9,10,1,4,5,2,3,6), as already suggested in the binary case in~\cite{CaldersG05}.  Since generators correspond to largest rectangles, i.e. on which the fewest minimal changes have been applied, if $c$ is not a generator, a generator $e$ associated to its equivalence class has already been generated, and $c$ is discarded. To check the existence of $e$, we look up in an auxiliary data-structure storing already extracted FIPG.  Precisely, if the data structure contains a FIPG $e$ with same support than candidate $c$, such that $e \sqsubseteq c$, $c$ is discarded, and the algorithm backtracks. Otherwise $c$ is declared as a FIPG and stored. We have experimented the \textit{MinIntChangeG} algorithm with two well-known and adapted data structures, a trie and a hashtable.

\section{Experiments}\label{exp}
We evaluate the performances of the algorithms designed in Java, namely \textit{MinIntChange}, \textit{MinIntChangeG-h} with auxiliary hashtable and \textit{MinIntChangeG-t} with auxiliary trie. Recalling that closed IS-itemsets and CIP are in 1-1-correspondence, we compare the performance for mining interordinal scaled data with the closed-itemset-mining algorithm \textit{LCMv2}~\cite{UnoKA04}. For studying the global redundancy effect of IS-itemset generators, we use the generator-mining-algorithm \textit{GrGrowth}~\cite{LiuLWH06}. Both implementations in C++ are available from the authors. All experiments are conducted on a 2.50Ghz machine with 16GB RAM running under Linux 2.6.18-92.e15. We choose dataset from the Bilkent repository\footnote{http://funapp.cs.bilkent.edu.tr/}, namely Bolts (BL), Basketball (BK) and Airport (AP), AP being worst case where each attribute value is different.

First experiments compare \textit{MinIntChange} for extracting FCIP and \textit{LCMv2} for extracting equivalent frequent closed IS-itemsets in Table~\ref{closedexec}. Second experiments consist in extracting frequent interval pattern generators (FIPG) with \textit{MinIntChange-h} and \mbox{\textit{MinIntChange-t}}. We also extract frequent itemset generators (FISG) in corresponding binary data after interordinal scaling with \textit{GrGrowth} for studying the global redundancy effect in Table~\ref{genexec}.
\begin{table}[h]\centering
\vspace{-0.2cm}
\begin{scriptsize}
\begin{tabular}{ccccc}
\hline\hline
Dataset  & minSupp & MinIntChange  & LCMv2& $|FCIP|$ \\ 
\hline\hline
BL & 80\% & $<50$ & $<50$ & 1,130  \\ 
&50\% & 252 &  \textbf{100} & 32,107 \\ 
&25\% & 1,215 & \textbf{1,060} & 171,192 \\ 
&10\% & \textbf{1,821} & 1,950 & 268975\\ 
&1 & \textbf{1,905} & 2,090 & 272,223 \\ 
\hline \hline
AP & 80\% & 4,595 & \textbf{1,470} &  346,741\\ 
&50\% & \textbf{143,939}  & 149,580 &  16,214,345 \\ 
&25\% & \textbf{413,805} & 899,180  & 58,373,631 \\ 
&10\% & \textbf{506,985} & 6,810,125 & 80,504,566\\
&1  & \textbf{517,548} & 6,813,591 & 82,467,124\\
\hline\hline 
\end{tabular} 
\end{scriptsize}
\vspace{-0.2cm}
\caption{Execution time for extracting FCIP (in ms).}
\vspace{-0.2cm}
\label{closedexec}
\end{table}

\begin{table*}\centering

\begin{scriptsize}
\begin{tabular}{cc||ccc||ccc||cc}
\hline\hline
Dataset & minSupp & GrGrowth & MinIntChangeG-h & MinIntChangeG-t & $|FIPG|$ & $|FISG|$ & $\frac{|FIPG|}{|FISG|}$ & $|FCIP|$ & $\frac{|FIPG|}{|FCIP|}$\\ 
\hline
BL &90\% & $<50$ &  $<50$ & $<50$ & 176    & 194     & 90\% & 112    & 1.57 \\ 
&80\% & $<50$ &  $<50$ & $<50$ & 1,952   & 2,823    & 69\% & 1,130   & 1.73 \\ 
&50\% & \textbf{150}   & 1,212   & 529  & 66,350 & 222,088 & 29\% & 32,107 & 2\\ 
&25\% & \textbf{3,432}  & 27,988  & 3,893 & 411,442 & 3,559,419 & 11\% & 171,192 & 2.4 \\ 
&1 & 123,564 & 438,214 & \textbf{24,141} & 1,165,824 & 69,646,301 & \textbf{1.6}\% & 272,223 & 4.3\\ 
\hline
BK & 90\% & \textbf{$<50$} & 1,268& 1,207 & 67,737 & 75,058 & 84\% & 48,847 & 1.3\\ 
&85\% & \textbf{4,565} & 26,154  & 12,139 & 554,956 & 799,574 & 69\% & 403,562 & 1.37 \\ 
&80\% & Untractable & 512,126  & \textbf{107,700} & 2,730,812 & NA & NA & 1,938,984 & 1.40 \\ 
\hline\hline 
\end{tabular} 
\vspace{-.1cm}
\caption{Execution time for extracting FIPG and global redundancy evaluation.}
\label{genexec}
\end{scriptsize}
\vspace{-.5cm}
\end{table*}

In both cases, using binary data is better when the minimal support is high (e.g. 90\%). For low supports, a critical issue, our algorithms deliver better execution times. Most importantly, the global redundancy effect discards the use of binary data, e.g. only $1.6\%$ of all FISG are actually FIPG in dataset BL. Finally, the algorithm \textit{MinIntChangeG-t} outperforms \textit{MinIntchangeG-h}. \textit{MinIntChangeG-t} however needs more memory since it stores each closed set of objects as a word in the trie, and to each word the list of associated FIPG.

It is very interesting to analyse the compression ability of closed interval patterns and generators. For that, we compare in each dataset the number of those patterns w.r.t. to all possible interval patterns. It gives the ratio of closed (generators) in the whole search space. In both cases, ratio varies between $10^{-7}$ and $10^{-9}$. This means that the volume of useful interval patterns, either closed or generators, is very low w.r.t. the set of all possible interval patterns, justifying our interest in equivalence classes for interval patterns.

\section{Conclusion}
We discussed the important problem of pattern discovery in numerical data with a new and original formalization of interval patterns. The classical FCA/itemset-mining settings are adapted accordingly: from a closure operator naturally rise the notions of equivalence classes, closed and generator patterns, and we designed corresponding algorithms. An appropriate semantics of interval patterns shows  from a theoretical (redundancy) and practical (computation times) points of view that mining equivalent binary data (encoding all possible intervals) is not acceptable. This is due to the fact that interval patterns are provided with a stronger partial ordering than IS-itemsets (classical set inclusion), hence pattern structures yield significantly less generators w.r.t. their semantics.

Dealing with interval patterns has applications in computational geometry, machine learning and data-mining, e.g.~\cite{BorosEGKM03} and references therein. It is indeed highly related to the actual problem of (maximal) $k$-boxes which corresponds to interval patterns (generators) with support $k$. When $k=0$, it corresponds to largest empty subspaces of the data. Our contribution to this field is the characterization of smaller subsets (closed and generators). 

In data-mining, closed patterns and their generators are crucial for extracting valid and informative association rules~\cite{BastideTPSL00}, while generators can be preferable to closed patterns following the minimum descriptions length principle for so-called itemset-based classifiers~\cite{LiLWPD06}. How these notions can be shifted to interval patterns is an original perspective of research rising questions concerning missing values, fault-tolerant patterns, and interestingness measures that are critical issues even in classical itemset mining: although the compression ability of closed interval patterns and generators is spectacular, the number of patterns remains too high for large datasets. However, bringing the problem of numerical pattern mining into well known settings in favor of these perspectives of research.

\bibliographystyle{named}

\end{document}